\documentclass[manuscript, nonacm]{acmart}
\usepackage[utf8]{inputenc}

\usepackage{microtype}
\usepackage{caption}
\usepackage{subcaption}
\usepackage{booktabs} 
\usepackage{amsmath}
\usepackage{amsthm}
\usepackage{amsfonts}       
\usepackage{mathtools}
\usepackage{nicefrac}
\usepackage{algorithm,algorithmic}
\usepackage{listings}
\newcommand{\RecSimNG}{{\textsc{RecSim NG}}}
\newcommand{\RecSim}{{\textsc{RecSim}}}

\newcommand{\inlinecode}[1]{{\small \texttt{{#1}}}}

\usepackage{accents}
\usepackage{amsfonts}
\usepackage{bm}
\usepackage{booktabs}
\usepackage{enumerate}
\usepackage{multirow}
\usepackage{pbox}

\usepackage{graphicx}
\usepackage{amsmath}
\usepackage{listings}

\usepackage{xcolor}

\usepackage{ragged2e}

\DeclareTextFontCommand{\emph}{\em}

\usepackage{xcolor}
\usepackage[backgroundcolor=white]{todonotes}

\newcommand{\commentout}[1]{}

\newcommand{\junk}[1]{}

\theoremstyle{definition}

\newcommand{\calS}{\mathcal{S}}

\definecolor{codegreen}{rgb}{0,0.6,0}
\definecolor{codegray}{rgb}{0.5,0.5,0.5}
\definecolor{codepurple}{rgb}{0.58,0,0.82}
\definecolor{backcolour}{rgb}{0.95,0.95,0.92}

\lstdefinestyle{mystyle}{
    backgroundcolor=\color{backcolour},   
    commentstyle=\color{codegreen},
    keywordstyle=\color{magenta},
    numberstyle=\tiny\color{codegray},
    stringstyle=\color{codepurple},
    basicstyle=\ttfamily\footnotesize,
    breakatwhitespace=false,         
    breaklines=true,                 
    captionpos=b,                    
    keepspaces=true,                 
    numbers=left,                    
    numbersep=5pt,                  
    showspaces=false,                
    showstringspaces=false,
    showtabs=false,                  
    tabsize=2
}

\newtheoremstyle{TheoremNum}%
    {\topsep}{\topsep}
    {\itshape}
    {}
    {\bfseries}
    {.}
    { }
    {\thmname{#1}\thmnote{ \bfseries #3}}
\theoremstyle{TheoremNum}

\title[RecSim NG]{RecSim NG: Toward Principled Uncertainty Modeling for Recommender Ecosystems}\titlenote{A shorter abstract of this work appeared as ``Demonstrating Principled Uncertainty Modeling for Recommender Ecosystems with RecSim NG,'' Demonstration paper at \emph{RecSys '20: Fourteenth ACM Conference on Recommender Systems (2020).}}

\author{M. Mladenov}
\authornote{Contact author.}
\affiliation{%
  \institution{Google Research}
  \streetaddress{1600 Amphitheatre Parkway}
  \city{Mountain View}
  \state{CA}
  \postcode{94043}
}
\email{mmladenov@google.com}

\author{C. Hsu}
\authornotemark[2]
\affiliation{%
  \institution{Google Research}
  \streetaddress{1600 Amphitheatre Parkway}
  \city{Mountain View}
  \state{CA}
  \postcode{94043}
}
\email{cwhsu@google.com}

\author{V. Jain}
\affiliation{%
  \institution{Google Research}
  \streetaddress{1600 Amphitheatre Parkway}
  \city{Mountain View}
  \state{CA}
  \postcode{94043}
}
\email{vihanjain@google.com}

\author{E. Ie}
\affiliation{%
  \institution{Google Research}
  \streetaddress{1600 Amphitheatre Parkway}
  \city{Mountain View}
  \state{CA}
  \postcode{94043}
}
\email{eugeneie@google.com}

\author{C. Colby}
\affiliation{%
  \institution{Google Research}
  \streetaddress{1600 Amphitheatre Parkway}
  \city{Mountain View}
  \state{CA}
  \postcode{94043}
}
\email{ccolby@google.com}

\author{N. Mayoraz}
\affiliation{%
  \institution{Google Research}
  \streetaddress{1600 Amphitheatre Parkway}
  \city{Mountain View}
  \state{CA}
  \postcode{94043}
}
\email{nmayoraz@google.com}

\author{H. Pham}
\affiliation{%
  \institution{Google Research}
  \streetaddress{1600 Amphitheatre Parkway}
  \city{Mountain View}
  \state{CA}
  \postcode{94043}
}
\email{huberpham@google.com}

\author{D. Tran}
\affiliation{%
  \institution{Google Research}
  \streetaddress{1600 Amphitheatre Parkway}
  \city{Mountain View}
  \state{CA}
  \postcode{94043}
}
\email{trandustin@google.com}

\author{I. Vendrov}
\affiliation{%
  \institution{Google Research}
  \streetaddress{1600 Amphitheatre Parkway}
  \city{Mountain View}
  \state{CA}
  \postcode{94043}
}
\email{ivendrov@google.com}

\author{C. Boutilier}
\affiliation{%
  \institution{Google Research}
  \streetaddress{1600 Amphitheatre Parkway}
  \city{Mountain View}
  \state{CA}
  \postcode{94043}
}
\email{cboutilier@google.com}

\begin{document}

\begin{abstract}
The development of recommender systems that optimize \emph{multi-turn interaction} with users, and model the interactions of different agents (e.g., users, content providers, vendors) in the recommender \emph{ecosystem} have drawn increasing attention in recent years.
Developing and training models and algorithms for such recommenders can be especially difficult using static datasets, which often fail to offer the types of counterfactual predictions needed to evaluate policies over extended horizons. To address this, we develop
\RecSimNG, a probabilistic platform for the simulation of multi-agent recommender systems. \RecSimNG\ is a scalable, modular, differentiable simulator implemented in Edward2 and TensorFlow. It offers: a powerful, general probabilistic programming language for agent-behavior specification; tools for probabilistic inference and latent-variable model learning, backed by automatic differentiation and tracing; and a TensorFlow-based runtime for running simulations on accelerated hardware. We describe \RecSimNG\ and illustrate how it can be used to create transparent, configurable, end-to-end models of a recommender ecosystem, complemented by a small set of simple use cases that demonstrate how \RecSimNG\ can help both researchers and practitioners easily develop and train novel algorithms for recommender systems.
\end{abstract}


\maketitle

\section{Introduction}
\label{sec:intro}

Recent years have seen increased emphasis, both in research and in practice, on recommender systems that are capable of sophisticated interaction with users. State-of-the-art recommender systems have moved beyond traditional models that simply present items and passively observe immediate user reactions (e.g., clicks, consumption, ratings, purchase, etc.), and have evolved to include systems capable of exploring user interests \cite{Li:2016:cfbandits,christakopoulou:2018:banditcir}, optimizing user engagement over multi-step horizons \cite{shani:jmlr05,facebook_horizon:2018,chen_etal:2018top,zhao_slateRL:recsys18,slateQ:ijcai19}, or engaging in natural language dialogue \cite{christakopoulou:2016:prefelicit,sun:2018:convrec}. Loosely speaking, such recommenders can often be viewed as engaging in multi-turn (or \emph{non-myopic}), cooperative exploration with the user to better serve the user's needs. We use the
term \emph{collaborative interactive recommenders (CIRs)} as a catch-all term for such systems.

The development of cutting-edge CIRs faces a number of challenges. In contrast to traditional myopic approaches to recommender systems, CIRs cannot generally be trained using \emph{static} data sets comprised of recommender actions and observed user (immediate) responses, even when organized into trajectories---predicting the impact of \emph{counterfactual actions} on user behavior is crucial. Moreover, once we model non-myopic user behavior, the \emph{interaction} between users (e.g., content consumers, potential customers) and providers (e.g., content creators or providers, vendors or retailers) takes on added importance in predicting recommender performance. Indeed, almost every practical recommender system lies at the heart of a complex, multi-agent 
\emph{ecosystem}. This amplifies the need for recommender methods to capture the long-term behavior of participants, as well as the (potentially strategic) interactions that emerge between them. Finally, because CIRs often adopt novel user interfaces or interaction modalities, there is often quite limited training data from which to learn how users respond to recommender actions within these new modalities.

Simulation models provide an effective way to tackle the challenges above. Models of user behavior---how users respond to novel recommender actions or interaction modes---and how it evolves over time can be used to assess the performance of a proposed CIR over long horizons. Moreover, when coupled with models of the behavior of \emph{other} entities that engage with the recommender platform (e.g., content providers), simulation provides an effective way of evaluating the \emph{dynamics} of the recommender ecosystem as a whole.

The use of simulation comes with its own challenges, of course, as we discuss below. To facilitate the development and study of recommendation models and algorithms in the complex environments characteristic of CIRs, we developed \RecSimNG, a configurable platform for both authoring and learning recommender system \emph{simulation environments}. Effective simulation can be used to evaluate existing recommender policies, or generate data to train new policies (in either a tightly coupled online fashion, or in batch mode). Just as progress in reinforcement learning (RL) research has been greatly accelerated by simulation \cite{bellemare:jair2013,castro18dopamine,OpenAI_Gym:arxiv16}, we believe that \RecSimNG\ can support the development of state-of-the-art CIRs and ecosystem-aware recommenders. Broadly, \RecSimNG\ provides a probabilistic programming framework that supports: (a) the natural, modular, ``causal'' specification of user and other agent behavior, and its dynamics, within CIR ecosystems; (b) the ability to learn the parameters of such models from data; and (c) a number of probabilistic inference techniques, beyond straightforward Monte Carlo simulation, for evaluating recommender algorithms.

More concretely, the main goals of \RecSimNG\ and our contributions embody the following principles:

\begin{itemize}
    \item \textbf{Models of agent behavior (e.g., users, content providers) across a \emph{sequence} of interactions with a CIR must reflect that agent's \emph{state} and how it evolves while engaging with the CIR.} \RecSimNG\ incorporates the probabilistic programming environment Edward2 \cite{tran2018simple} to specify common design patterns for user state and behaviors (e.g., user preferences, satisfaction, choice and consumption behavior). \RecSimNG\ also allows the natural expression of causal generative models of user (and other participant) behavior and utility, including user state, choice, response and engagement models, and user-state transition dynamics. These models are specified as a composable set of \emph{dynamic Bayesian networks (DBNs)} \cite{dean-kanazawa:ci89,lerner_hybridDBN:uai02} which we organize in an object-oriented fashion \cite{pfeffer:uai97} using three main concepts:
    \emph{entities} (e.g., users, recommenders), \emph{behaviors} (e.g., state transitions, user choices) and \emph{stories} (e.g., user-system interaction details).

    \item \textbf{Models of user state must support \emph{latent} or unobservable state variables.} CIRs must often engage in (implicit or explicit) latent state estimation from various observable behaviors to build internal models of user preferences, user psychological state (e.g., satisfaction, frustration), and other exogenous environmental factors to compute effective (long-term) recommendation strategies. \RecSimNG\ allows the flexible specification of both observable and latent aspects of a user's or agent's state so that practitioners can experiment with the state-estimation capabilities of different model architectures.
    
    \item \textbf{Behavior models require flexibility in their model structure and probabilistic inference methods.}
    \RecSimNG\ allows one to impose as much (or as little) structure on behavior models as needed.
     For example, in psychometric models of user choice, it is quite natural to specify structural priors, biases and parametrizations, while the precise parameters must usually be estimated from behavioral data. \RecSimNG\ supports the flexibility of imposing model structure and learning of model parameters from data, or learning the structure itself. Because many realistic models have latent factors, model learning requires sophisticated probabilistic inference, beyond the usual Monte Carlo rollouts embodied by most simulation environments. \RecSimNG\ supports latent variable inference, including several Markov-chain Monte Carlo (MCMC) and variational methods.
     
    \item \textbf{Ecosystem modeling is critical in complex CIRs, including the modeling of: incentives of individual agents that drive their behaviors; variable observability criteria for pairs/groups of agents; and the \emph{interaction} between agents as mediated by the CIR.} The use of entities, stories and behaviors by \RecSimNG\ allows the natural specification of the interaction between agents in the system, while making it easy to uncover any independence that exists. Moreover, scalability is critical, especially in CIRs with large populations. \RecSimNG\ provides scalable TensorFlow-based execution to support several forms of posterior inference beyond standard MC rollouts.

    \item \textbf{Modular modeling of agent behavior provides valuable flexibility in generating data to test the capabilities of novel CIR algorithms, models and modes of interaction.} The agent-behavior abstractions offered by \RecSimNG---based on probabilistic graphical models and adopting an object-orientation---encourages the decomposition of agent behaviors into natural, reusable components that can be used in a variety of different environments. This ensures relatively stable behaviors that facilitate the comparison of different recommendation strategies and interaction modes. For example, one can readily compare the performance of two CIRs in the same domain that use slightly different interfaces (e.g., one GUI-based, the other voice-based) on the same ``population'' of simulated users by using the same user latent-state, state-transition and choice models, while changing only the user response model to accommodate the different interfaces.
\end{itemize}

In this work, we primarily view simulation as a tool to explore, evaluate, and compare different recommender models, algorithms, and strategies. While the so-called ``sim2real'' perspective is valuable, we focus on simulations that reflect \emph{particular} phenomena of interest---specific aspects of user behavior or ecosystem interaction arising in live systems---to allow the controlled evaluation of recommender methods at suitable levels of abstraction. We illustrate this below with several specific use cases. \RecSimNG\ should be equally valuable to researchers (e.g., as an aid to reproducibility and model sharing) and practitioners (e.g., to support rapid model refinement and evaluation prior to training in a live system). That said, \RecSimNG\ may also serve as a valuable tool to generate training data for live recommender systems.

The remainder of the paper is organized as follows. We discuss related work in Sec.~\ref{sec:related}. We describe \RecSimNG\ in Sec.~\ref{sec:recsim} in stages. We first explicate the modeling language in general terms (Sec.~\ref{sec:overview}), then detail its use of entities, behaviors and stories more concretely using a simple, stylized example (Sec.~\ref{sec:examplemodel}). We next illustrate the use of some reusable behavioral building blocks from the \RecSimNG\ library in a slightly more elaborate example (Sec.~\ref{sec:library}). We also describe \RecSimNG's scalable execution in TensorFlow (Sec.~\ref{sec:tensorflow}) and probabilistic inference and learning capabilities (Sec.~\ref{sec:learning}). In Sec.~\ref{sec:usecases}, we present three use cases that demonstrate \RecSimNG's capabilities and provide simple but suggestive empirical results.

The \RecSimNG\ library is available for download,\footnote{See \texttt{https://github.com/google-research/recsim\_ng}.} and more detailed exposition of it capabilities and its usage can be found in the tutorials offered there. Some of the details of the examples and use cases presented in the sequel are left unstated, but full details (including models, algorithms and their specific parametrizations) can be found in the library.

\section{Related Work}
\label{sec:related}

\RecSimNG---both the underlying framework and motivating use cases illustrating its capabilities---draw inspiration from multiple research areas. These include work on classical recommender systems, reinforcement learning, sequential models in recommender systems, conversational recommenders, multi-agent modeling, and probabilistic programming, to name a few. A detailed review of each of these areas is beyond the scope of this expository overview, hence we briefly discuss just a few key connections and influences.

Perhaps the dominant ``classical'' approach to recommender systems is \emph{collaborative filtering (CF)}, which includes early matrix factorization models \cite{grouplens:cacm97,breese-cf:uai98,salakhutdinov-mnih:nips07} and more recent neural CF approaches \cite{covington:recsys16,he_etal:www17,yangEtAl:www20} to predict users' item preferences. Increasingly, recommenders predict more detailed user responses to recommendations, such as click-through-rates (CTR) or engagement (e.g., dwell/watch/listen time, purchase behavior, etc.) \cite{vandenoord:nips13,covington:recsys16}.
We use CF-based embeddings and user response models in several of the \RecSimNG\ use cases discussed in the sequel..

Modeling sequential user behavior in using RL to optimize recommender strategies has long been considered important \cite{shani:jmlr05,taghipour:recsys07}. With the advent of deep RL, such models have attracted considerable attention over the past several years \cite{choi2018reinforcement,facebook_horizon:2018,chen_etal:2018top,zhao_slateRL:recsys18,slateQ:ijcai19}. Indeed, sequential behavior modeling and RL-based optimization is a key motivation for \RecSimNG. The difficulty of evaluating RL methods using static datasets \cite{harper16:movielens}, due to the impact of counterfactual actions on user trajectories, is a primary reason to use simulation
\cite{rohde2018recogym,facebook_horizon:2018,recsim:arxiv19}.\footnote{Indeed, simulation is a critical component or RL research as we elaborate below.} We illustrate a simple RL use case below (drawing also on elements from bandit models in recommenders, e.g., \cite{Li:2016:cfbandits,christakopoulou:2018:banditcir}).
We also use \RecSimNG\ to model (and learn) aspects of latent user state. Explicit latent variable models have been explored on occasion in RSs
(e.g., \cite{sahoo:mis2012,gentile:icml14,maillard:icml14,bonner_rodhe_latent_recs:arxiv19}). However, the use of recurrent models has been method of choice is recent models of sequential user behavior \cite{ahmed_smola_et_al:wsdm17,tan_rnn_recommender:2016}.

Ecosystem modeling in recommenders is somewhat uncommon, though it has some connections to the study of \emph{fairness} in ML.
Studies of group fairness in ranking, for example, often by ``regularize'' standard ranking metrics like NDCG
to encourage parity of the ranks across demographic groups \cite{Yang2016MeasuringFI,Zehlike2017FAIRAF}. Of special relevance is recent work that attempts to fairly assign users/exposure to content providers \cite{SinghJoachims2018,Biega}. Also related is a recent game-theoretic
model of recommender systems whose providers act strategically---by making available
or withholding content---to maximize
their own user engagement on the platform
\cite{benporat_etal:nips18,benporat_etal:aaai19}. Finally, recent work has explored the impact of a recommender's policies on content provider behavior, how this impacts long-term user utility, and how to optimize policies under such conditions to maximize social welfare over extended horizons \cite{mladenov_etal:icml20}. We adopt a simplified variant of this latter model
in one of our illustrative use cases in Sec.~\ref{sec:usecases}.

As a simulation platform, \RecSimNG\ shares much with recent systems used in RL and recommender systems research. Simulation has played an critical role in the evaluation of RL methods in recent years. Many of these platforms, including 
ALE \cite{bellemare:jair2013}, Dopamine \cite{castro18dopamine}, and OpenAI Gym \cite{OpenAI_Gym:arxiv16} offer of a collection of environments against which RL algorithms can be benchmarked and compared. These platforms often provide a set of well-known RL algorithms for comparison. Our work shares some of the same motivations, but differs in that we focus on allowing the \emph{direct authoring and learning} of environments to push development of algorithms that handle new domain characteristics rather than on benchmarking \emph{per se}.
ELF \cite{ELF:nips2017} is similar to \RecSimNG\ in this respect, as it allows configuration of real-time strategy games to support the development of new RL methods, overcoming some of the challenges of doing research with commercial games (e.g., by allowing access to internal game state).

Simulation has recently found direct use in recommenders as well. Three platforms are especially  related to \RecSimNG. \citet{yao_halpern_etal:fates20} use extensive simulation studies to examine the impact of recommender policies on long-term user behavior, and specifically to disentangle the influence of the recommender itself from that of a user's own preferences on this behavior. While not a configurable platform like \RecSimNG, the models developed in that work share much with some of the model structures \RecSimNG\ is intended to support.
\citet{rohde2018recogym} propose RecoGym, a configurable recommender systems simulation environment. It allows the study of sequential user interaction that combines organic navigation with intermittent recommendation (or ads). RecoGym, however, does not allow configuration of 
user state transitions, but instead allows bandit-style feedback.
Finnaly, we note that \RecSimNG\ is a major extension and reconception of the earlier \RecSim\ platform developed by \citet{recsim:arxiv19}. Like RecoGym, \RecSim\ is configurable and is well-suited to the study of sequential user interaction. It also allows the relatively straightforward configuration of user state dynamics. \RecSimNG\ shares much with these platforms, but offers the ease, generality and flexibility of probabilistic programming, general probabilistic inference (not just trajectory generation), and scalable execution in TensorFlow. 

Rule-based and learning-focused environments have been used for goal-oriented dialog agents \cite{wei_airdialogue:emnlp18,peng:2018:deepdynaq} and interactive search \cite{liu_etal:ijcai19}. Generative adversarial networks have been
used to generate virtual users for high-fidelity recommender environments to support
learning policies that can be transferred to real systems \cite{shi_taobao:aaai19,zhao_RLSim:arxiv19}.
Multi-agent simulation has a long history, and has recently found use for modeling agent interactions in economic \cite{2004.13332}, social media \cite{2004.05363} and social network settings \cite{blythe:MMAS18,tregubov:MABS20}. None of these models support the authoring of general interaction models.

As a probabilistic programming framework, \RecSimNG\ builds on the area of deep probabilistic programming, enabling probabilistic models with deep neural networks, hardware accelerators, and composable inference methods \citep{tran2017deep,bingham2019pyro,tran2019bayesian}. \RecSimNG\ proposes new abstractions for user modeling, multi-agent modeling and ecosystem interaction that are well-suited to CIRs. There has been prior work on multi-agent probabilistic systems \citep{goodman2014dippl,evans2017modeling}, typically rooted in cognitive science. These investigations usually involve only a handful of agents and small (if any) datasets. They also do not examine user-item ecosystems or item providers nor do they offer support for flexible probabilistic inference methods, neural network models, etc. 

\section{\RecSimNG\ Modeling Environment}
\label{sec:recsim}

In this section, we outline the \RecSimNG\ modeling formalism. We start with a low-level formal description of the \RecSimNG\ model semantics. We then present the high-level abstractions of behaviors and entities, give an overview of hierarchical agent modeling and the \RecSimNG\ behavior modeling library, and conclude with a brief introduction to probabilistic reasoning in \RecSimNG.

\subsection{Modeling Formalism}
\label{sec:overview}

Abstractly, a \RecSimNG\ \emph{model} (also called a \RecSimNG\ "simulation") represents a \emph{Markovian stochastic process}. We assume some state space $\calS$  over which a Markov process gives rise to \emph{trajectories}, that is, sequences of states from $\calS$ of length $n$, $(s_i \in {\calS})_{i=1}^n$, such that \[p(s_0, s_1, \ldots, s_{n-1}) = p_0(s_0)\prod_{i}T(s_{i}, s_{i+1}),\]
where $p_0$ is an initial state density, and $T$ is a transition kernel. We can think of $\calS$, for example, as the state of the recommender ecosystem, which itself might be the collection of states of all the individual agents in it.

In most cases, $s \in \calS$ is not just some vector in Euclidean space. Instead, $\calS$ is a highly structured object consisting of sub-components (e.g., the states of the individual agents) which may evolve autonomously, or through interactions with each other. 
In \RecSimNG\, we assume that the state space factors as the Cartesian product ${\calS} = \times_i S^i$, where each $S^i$ is a {\bf component} of the state space. For example, each component might reflect the state of some agent, or some random variable in the recommender environment (say, the weather or traffic conditions, or an emerging news trend). A factorization of the state space allows us to specify the initial state and transition kernel in a factored way as well. To do so, we use the language of \emph{Dynamic Bayesian Networks (DBNs)} \cite{dean-kanazawa:ci89}.

Suppose that $\Gamma, \Gamma_{-1}$ are two (possibly different) directed acyclic graphs (DAGs) whose nodes are the components of the state space. We then assume that 
$$T(s_t, s_{t-1}) = \prod_i T^i(s_t^i | s^{\operatorname{Pa}_{\Gamma}(i)}_t, s_{t-1}^{{\operatorname{Pa}_{\Gamma_{-1}}(i)}}),$$
that is, the state of the component $i$ depends on the state of its parents in $\Gamma$ (its \emph{intra-slice dependencies}), as well as the preceding state of its parents in $\Gamma_{-1}$ (its \emph{inter-slice dependencies}). The components $T^i$ of the transition kernel are termed \emph{conditional probability densities} (CPDs) or sometimes \emph{factors}.
We assume that the initial state density can be factored similarly, but the graphs may be different (we will not spell this out explicitly). 

It is important to note that the notion of Markovianness employed above is defined exclusively from the point of view of \emph{the simulation runtime}, \emph{not} from the perspective of any individual agent or action (e.g., user, content provider, recommender platform) in the simulation. This implies that the simulator must be able to observe all random variables---for example, the simulator must have  access to the state of all agents. This does \emph{not} impact the ability of \RecSimNG\ to simulate partially observable problems (e.g., partially observable MDPs where users have latent state that is not directly observable by the recommender). In such scenarios, only \emph{part} of the simulation state is communicated to a particular decision-making agent, meaning that its own state CPD can only depend on a specific subset of the simulation variables.

A \RecSimNG\ model is thus a specification of some number of component random variables, their initial distributions, their transition kernels, and the structure of the inter- and intra-slice dependencies. We illustrate this with a simple example of how these are implemented in \RecSimNG\ using a deterministic simulation that counts to $n$.
\begin{lstlisting}[language=Python]
# Declare a component variable, its name and state space.
count_var = Variable(name="count", spec=ValueSpec(n=FieldSpec()))
# Define the variable's initial state, in this case deterministic.
def count_init() -> Value:
    return Value(n=0)
# Define the variable's kernel.
def count_next(previous_value: Value) -> Value:
    n_prev = previous_value.get("n")
    return Value(n=n_prev + 1)
# Finally bind the initial state distribution and kernel to the variable.
count_var.initial_value = variable.value(count_init)
# We declare an inter-slice dependency on the variable's previous value.
count_var.value = variable.value(count_next, (count_var.previous,))
\end{lstlisting}

The snippet above declares a single state component $s^0$ with name \inlinecode{"count"} represented by the Python variable \inlinecode{count\_var}. Its initial distribution $p_0$ is established by binding \inlinecode{count\_var.initial\_value} to a function that samples from $p_0\left(s_0^0\left|s^{\operatorname{Pa}_{\Gamma_0}(0)}_0\right.\right)$ (note that $s^{\operatorname{Pa}_{\Gamma_0}(0)}_0$ happens to be the empty set in this case), and the transition kernel $T$ is established by binding \inlinecode{count\_var.value} to a function sampling from $T^i(s_t^0 | s^{\operatorname{Pa}_{\Gamma}(0)}_t, s_{t-1}^{{\operatorname{Pa}_{\Gamma_{-1}}(0)}})$, where $s^{\operatorname{Pa}_{\Gamma}(0)}_t=\emptyset$ and $s^{\operatorname{Pa}_{\Gamma_{-1}}(0)}_t = \{0\}$. The set of parents of a component random variable is declared by the list of components in the second argument of \inlinecode{variable.value}, where the \inlinecode{.previous} property indicates a $\Gamma^{-1}$ dependency (for $\Gamma$ dependencies, the variables are used as is). All data-generating functions that interact with the \RecSimNG\ runtime return a \inlinecode{Value} object, which is essentially a structured dictionary.

Once fully defined, $\mathtt{count\_var}$ can be simulated by the \RecSimNG\ runtime to generate trajectories or simply return the variable's state after $n$ steps of simulation. We discuss dependencies in somewhat more detail in Sec.~\ref{sec:examplemodel}, but refer to the online tutorials for a more complete treatment.\footnote{See \texttt{https://github.com/google-research/recsim\_ng}.}

\subsection{Scalable Behavioral Modeling in \RecSimNG}
\label{sec:examplemodel}

Component random variables together with their distributions and dependencies are the ``assembly language`` of \RecSimNG. The $\mathtt{Variable}$ API is intentionally kept simple in order to avoid unnecessarily restricting the range of applications that can be implemented. There are, however, several reasons not to rely on variables alone for complex recommender system simulations. First, even a simple recommender ecosystem model may involve millions of random variables per time-step, such as user choices, evolving preferences, stochastic recommendation policy decisions, and so on. Modeling these as individual random variables within a ``flat'' DBN would be extremely tedious. Second, expressing agent dynamics exclusively through kernels is somewhat contrary to how we tend to think of agents. Intuitively, an agent is an actor with specific properties (goals, preferences, etc.) that engages in behaviors to achieve some desired outcome. For example, a content provider may produce content that aims to maximize user engagement subject to budget/time constraints; an advertiser may bid on advertising opportunities to maximize conversion rates; or a user may issues a sequence of queries and consume particular recommended content that helps meet some informational or entertainment goal.

In an agent-based simulation, there are two common sources of structure that can be harnessed to make complex models easier to implement:

\begin{itemize}
\item \emph{Parameter sharing/encapsulation}: An agent may have multiple behaviors such as ``choose,`` ``observe,`` ``update state,`` etc. All of these behaviors might be influenced by a common set of parameters, such as the agent's preferences, its risk sensitivity, its overall satisfaction, etc. This fits nicely within the object-oriented paradigm in which the various behaviors are expressed as methods and the shared parameters are encapsulated properties of the object. (In fact, agent-based modeling is often used to justify object-oriented programming.) 
\item \emph{Behavioral generalization}: While every agent in a population is unique to some degree, not everything they do is idiosyncratic. In fact, most behavioral models tend to be expressed in terms of a population ``template'' that abstracts agent individuality by reducing it to some small set of agent-specific, ``personalized'' parameter values. For example, a Luce-Shephard or multinomial logit (MNL) model for \emph{user choice} (i.e., a model which specifies the probability with which a user selects a specific item from a slate of recommended items) posits that an agent $a$ will choose an item $i$ with probability proportional to $e^{f_{\mathit{aff}}(\mathbf{a},\mathbf{i})}$, where $\mathbf{a}$ is the agent's feature vector, $\mathbf{i}$ is a feature vector characterizing the item, and $f_{\mathit{aff}}$ is an affinity function.
While the agent features are personalized to each agent, the affinity model and the choice distribution family (softmax/MNL in this case) are common to the entire population. Templatized models of this form offer two critical benefits: (a) they allow us to define large populations of agents very concisely; and (b) they can make heavy use of the accelerated computation available on modern hardware (through, for example, highly parallel hardware architectures).
\end{itemize}

\vskip 2mm
\noindent
\textbf{Entities and Behaviors.} In \RecSimNG, the two types of structure above are harnessed through the ``Entity'' pattern. A \RecSimNG\ Entity is a class that models the \emph{parameters and behaviors} of an entire population of agents, making use of batched execution as much as possible. 

The following example shows a partially implemented Entity prototype for a user model. This user model reflects a configurable user population of flexible size (as reflected by the \inlinecode{num\_users} parameters). The state of each user in this population is a $d$-dimensional \emph{interest vector} intended to represent a user's affinity for specific content items. Each user has three Behaviors, which both influence and are influenced by the user's state. The \inlinecode{initial\_state} behavior stochastically generates user initial state values. The \inlinecode{response} behavior is intended to model how the user responds to a slate of recommendations given its state (e.g., the probability of clicking, rating, commenting, etc.), while \inlinecode{next\_state} dictates how the user state evolves as a function of the previous state and the outcome of the current response. 

In the following snippet, the \inlinecode{initial\_state} state behavior is implemented to draw the user's initial interest vector from a $d$-dimensional unit normal distribution. Note that this behavior initializes the states of \emph{all users} in the entire population, as opposed to that of a single user. (We illustrate examples of the \inlinecode{response} and \inlinecode{next\_state} behaviors in the following subsection.)

\begin{lstlisting}[language=Python]
  class User(entity.Entity):
    def __init__(self, num_users, parameters) -> None:
      super().__init__(name="MyUserModel")
      self._parameters = parameters
      self._num_users = num_users

    def initial_state(self) -> Value:
      return Value(interest=ed.Normal(loc=tf.zeros(self._num_users, 
      self._parameters["pref_dimension"]), scale=self._parameters["initial_pref_scale"])))

    def next_state(self, old_state: Value, response: Value) -> Value:
      pass

    def response(self, state: Value, recommended_slate: Value) -> Value:
      pass
\end{lstlisting} 

A \RecSimNG\ entity owns several component random variables and declares their domains. Each behavior takes the values of some random variables as input and then returns the values of other random variables. The \inlinecode{Value} is a mapping of random variables.
The domains of random variables are declared using OpenAI Gym spaces \cite{OpenAI_Gym:arxiv16} in \inlinecode{specs()}, specifying the variable's shape (dimensions) and type (discrete, continuous). 
In the above example, \inlinecode{item} is a $d$-dimensional random variable but the others are just $1$-dimensional random variables. The resulting \inlinecode{ValueSpec} is a mapping of random variable domains.
Structured data types such as tuples, dictionaries, and nested compositions thereof are also supported.

\vskip 2mm
\noindent
\textbf{Stories.}
A story is the top-level code responsible for creating fully-defined component random variables. The function of a story is essentially to define a family of parametrized simulations by ingesting a set of user-defined parameters (such as number of users, model parameters, entity constructors and so on), creating all entities, component random variables, binding the component random variables to the entities' behaviors, and outputting the fully defined random variables, which can then be passed to the runtime for simulation.  Consider the following example.

\begin{lstlisting}[language=Python]
  def simple_user_recommender_story(num_users, user_model_ctor, recommender_agent_ctor):
    # Create entities.
    user_model = user_model_ctor(num_users)
    recommender_agent = recommender_agent_ctor()

    # Define random variables.
    user_state = Variable(name="user state", ...)
    response = Variable(name="user response", ...)
    rec_slate = Variable(name="slates", ...)

    # Bind random variables to behaviors.
    # CPDs at t=0.
    user_state.initial_value = variable.value(user_model.initial_state)
    rec_slate.initial_value = variable.value(recommender_agent.slate)
    user_response.initial_value = variable.value(user_model.response, (user_state, rec_slate))
    # Transition kernels.
    rec_slate.value = variable.value(recommender_agent.slate)
    response.value = variable.value(user_model.response, (user_state.previous, rec_slate))
    user_state.value = variable.value(
        user_model.next_state, (user_state.previous, user_response))
    return [rec_slate, user_response, user_state]
\end{lstlisting}

The use of stories is not strictly required for creating \RecSimNG\ simulations, however, it has two major benefits. First, it condenses the information flow between behaviors into a single function, and second, it allows us to templetize the creation of a parametrized suite of simulation scenarios. For example, we could create a suite of experiments testing how a given recommender deals with populations of varying sizes by running the story against multiple values of the \inlinecode{num\_users} parameter. Alternatively, we could test different pairs of recommender agents and user models by swapping out \inlinecode{user\_model\_ctor} and \inlinecode{recommender\_agent\_ctor}. 

Finally, the \RecSimNG\ \emph{learning APIs} offer some additional convenience when using stories. For example,

\begin{lstlisting}[language=Python]
component_rvs, trainable_variables = story_with_trainable_variables(lambda: simple_user_recommender_story(...))
\end{lstlisting}
captures all trainable TensorFlow variables created during the execution of the story.

\subsection{The \RecSimNG\ Behavioral Modeling Library}
 \label{sec:library}
 
While the entity/behavior/story pattern removes considerable complexity from the implementation of large-scale stochastic simulations, writing population-level stochastic simulations in TensorFlow can still be challenging in certain situations. Inspired by the success of compositional deep learning frameworks such as Keras,\footnote{See \texttt{https://keras.io}.}
\RecSimNG\ provides additional avenues for complexity reduction in the form of a curated library of \emph{behavioral building blocks}, which provide highly vectorized implementations of common elements in agent-based models. These are, in a sense, \emph{behavioral layers}. We provide a brief overview of the \RecSimNG\ library and illustrate the compositional APIs through which these building blocks are used.

\vspace*{2mm}
\noindent
\textbf{Choice models:}
Agent \emph{choice} is perhaps the most fundamental behavior in modeling recommender systems. Users often choose an item for consumption from a slate of recommended items (or abstain from choice); content providers might choose what type of content to offer at different points in time; and the policy of the recommender system itself involves the choice of a slate of items to present to users at different times. The totality of these choices jointly determines the satisfaction and behavior of all participants in the ecosystem.

The \RecSimNG\ modeling library currently supports the following modeling paradigm: when an agent (e.g., a user) is presented with a choice situation consisting of a slate of items, the agent's affinity for each item is calculated using an \emph{affinity model}.  An affinity model ingests a set of item features, as well as (optional) side information (e.g., user or environment context, user state) and outputs a vector of scores, one for each item on the slate (and optionally a ``no choice'' score). For instance, affinity for item $i$ might simply be the negative Euclidean distance between $i$ and the agent's \emph{target item} in some embedding space (here the target item would be part of the agent's state). The affinities are then passed as parameters to some choice distribution, such as a greedy, MNL, cascade or Plackett-Luce model, which samples zero or more chosen items. 

\vspace*{2mm}
\noindent
\textbf{State models:} Understanding the impact of recommendations on user behavior and recommender engagement metrics, especially over long horizons, requires understanding the user's \emph{state}. While observable features such as demographics are certainly valuable, many interesting user properties are \emph{latent}, i.e., not directly observable by the recommender system. Many latent state features are reasonably static---for instance, general item preferences and interests, or stable psychological or personality characteristics like curiosity, patience, etc. Others are transient and may exhibit interesting dynamics---for example, the user's current task, the current context (e.g., the user's companions in the moment, the activity the user is engaged in), or mental state (e.g., satisfaction, frustration).

\RecSimNG\ offers three categories of state models which can be applied to specific state variables:
\begin{itemize}
    \item Static: the state is sampled once at $t=0$ and then kept constant.
    \item Dynamic: the state evolves dynamically according to some (controllable) transition kernel. 
    \item Estimators/belief states: these aim to summarize a sequence of observations, compiling them into sufficient statistics that reflect an agent's belief about some unobserved state variable. These are typically used to represent the recommender's beliefs about, say, user state, but could also be used to model a user's state of knowledge about available content.
\end{itemize}
Static state models include various ways of sampling $N$-dimensional vectors from some structured distribution (e.g., a Gaussian mixture model, or hierarchical sampling of points from a set of hard clusters). Dynamic state models include finite-state controlled Markov chains, linear-Gaussian dynamical systems, as well as various recurrent neural network cells. Estimation models encompass methods like Kalman filters, direct posterior calculation for finite distributions, and simple finite-observation history arrays.

\vspace*{2mm}
\noindent
\textbf{Algorithmic primitives:} \RecSimNG\ offers implementations of a variety of common recommendation system algorithms. This includes a fairly extensive collection of bandit algorithms, including algorithms for the standard multi-armed bandit setting as well as various generalizations and extensions (e.g., linear and general contextual bandits). These methods cover many well-established algorithmic paradigms (e.g., optimism/UCB, posterior sampling) as well as black-box trainable RNN agents. These algorithms can be used in stand-alone fashion, or as part of more complex policies. They can even be used to implement non-recommender agents; e.g. a user's exploration behavior can be modeled as a multi-armed bandit.

\vspace*{2mm}
\noindent
\textbf{Prototypes:} The \RecSimNG\ library contains abstract classes for some commonly used types of agents (users, recommender, content providers), to serve as guidance for implementing custom classes. Inheriting from these classes also ensures that the custom implementations can be used within RecSim's existing recommender stories.

Note that not all of the above model types will be available in the initial release version of \RecSimNG. Moreover, this is not an exhaustive list---the library will grow as new applications are introduced.

\vspace*{2mm}
\noindent
\textbf{Illustration:} We conclude this section with an example of how these building blocks are used in the construction of Entities. We revisit the user-model Entity example from Sec.~\ref{sec:examplemodel}; but instead of manually implementing all behaviors, we use \RecSimNG\ building blocks. Specifically, we implement the following model:
\begin{itemize}

    \item The user's state consists of a $d$-dimensional interest vector $S$. Its initial value is distributed according to a $d$-dimensional unit normal distribution.
    
    \item At each time step, the user is presented with a slate of items $(1,\ldots,k)$. Each item $i$ is represented by a $d$-dimensional feature vector $F_i$, and a \emph{quality} scalar $q_i\in [-1, \ldots, 1]$. The user must chose one of the $k$ items for consumption. The probability of choosing any particular item is proportional to $\exp{-||F_i - S||}$. We note that quality $q_i$ plays no direct role in the user's choice---we assume that $q_i$ is only observed by the user if/once the item is chosen and consumed.
    
    \item Having consumed an item, the user's interest evolves as $S^t = S^{t-1} + \lambda q_i (F_i-S^{t-1}) + \xi$, where $\lambda$ is a sensitivity parameter and $\xi$ is Gaussian noise. Notice that if the consumed item has positive quality, the user's interest vector moves toward the item vector, and it moves away if the item has negative quality.
    
\end{itemize}

We map these elements to the \RecSimNG\ model library as follows. For interest dynamics, we use a \\
\inlinecode{ControlledLinearGaussianStateModel}, with an identity transition matrix and $\lambda I$ as a control matrix. This model takes $q_i (F_i-S^{t-1})$ as its control input. To implement user choice, we use a \inlinecode{TargetPointSimilarityModel} to compute the negative Euclidean distance of the user's interest to all items in the slate, then pass these affinities to a \inlinecode{MultinomialLogitChoiceModel}, which samples the chosen item. An abridged implementation is given below. Note that all of these building blocks support batch execution, so this code \emph{simultaneously} simulates the entire population of users. 
\begin{lstlisting}[language=Python]
  class User(entity.Entity):
    def __init__(self, num_users, parameters) -> None:
      super().__init__(name="MyUserModel")
      self._parameters = parameters
      self._num_users = num_users
      self._affinity_model = choice_lib.affinities.TargetPointSimilatrity(
       ..., similarity_type='negative_euclidean')
      self._selector = choice_lib.selectors.MultinomialLogitChoiceModel(...)
      self._interest_model = state_lib.dynamic.ControlledLinearScaledGaussianStateModel(...)

    def initial_state(self) -> Value:
      initial_interest = self._interest_model.initial_state()
      return initial_interest.prefixed_with("interest").union(Value(satisfaction=ed.Deterministic(loc=tf.ones(self._num_users))))

    def next_state(self, old_state: Value, response: Value) -> Value:
        chosen_docs = response.get('choice')
        chosen_doc_features = selector_lib.get_chosen(slate_docs, chosen_docs)
        # Calculate utilities.
        user_interests = previous_state.get('interest.state')
        doc_features = chosen_doc_features.get('doc_features')
        # User interests are increased/decreased towards the consumed document's
        # topic proportional to the document quality.
        direction = tf.expand_dims(
            chosen_doc_features.get('doc_quality'), axis=-1) * (
                doc_features - user_interests)
        next_interest = self._interest_model.next_state(
            previous_state.get('interest'),
            Value(input=direction))
        return next_interest.prefixed_with('interest')

    def response(self, state: Value, recommended_slate: Value) -> Value:
      affinities = self._affinity_model.affinities(
        state.get('interest.state'),
        recommended_slate.get('doc_features')).get('affinities')
      choice = self._selector.choice(affinities)
    return choice
\end{lstlisting} 
In \RecSimNG\ , model building follows a hierarchical pattern---building blocks are also entities owned by the ``parent'' entity. The building block entities output the result of the behaviors they implement, which can then be added to the state of the parent entity by means of the \inlinecode{Value.union} method. Additionally, the ``Value'' object supports (tree-based) hierarchical indexing. For example, if \inlinecode{user\_model.\_satisfaction\_model.next\_state} outputs \inlinecode{Value(state=...)} and \inlinecode{user\_model.\_interest\_model.next\_state} outputs \inlinecode{Value(state=...)}, then we can compose those with prefixes, as in: 
\begin{lstlisting}[language=Python]
interest_next_state = self._interest_model.next_state(interest_args)
satisfaction_next_state = self._satisfaction_model.next_state(satisfaction_args)
result = interest_next_state.prefixed_with('interest').union(satisfaction_next_state.prefixed_with('satisfaction'))
\end{lstlisting}
to create a \inlinecode{Value} object with keys \inlinecode{'interest.state'} and \inlinecode{'satisfaction.state'}.

\subsection{Runtimes and Model Execution}
\label{sec:tensorflow}

Once a set of component random variables has been fully defined, it is passed to a \RecSimNG\ runtime, which generates trajectories from the stochastic process. For example:
\begin{lstlisting}[language=Python]
model_dbn = network_lib.Network(list_of_variables)
runtime = runtime_ctor(network=model_dbn)
trajectory = runtime.trajectory(trajectory_length)
\end{lstlisting}
\RecSimNG\ features a modular runtime architecture, in which different types of runtimes can be implemented. A runtime is essentially an implementation of a sampler for a DBN. \RecSimNG\ offers a pure Python runtime, which implements direct ancestral sampling in a Python for-loop. A more sophisticated runtime is the TensorFlow runtime, which wraps the sampling process in a TensorFlow \inlinecode{scan} operator. The TensorFlow runtime allows the entire simulation to be compiled into a single TensorFlow graph, which confers some significant benefits, among them:
\begin{itemize}
    \item The simulation will take advantage of all optimizations offered by TensorFlow's AutoGraph compiler, including XLA (accelerated linear algebra) if available.
    \item The simulation will automatically utilize all available CPU or GPU cores on the host machine without any additional effort required by the \RecSimNG\ developer.
    \item The simulation graph can execute on specialized hardware such as Tensor Processing Units (TPUs).
\end{itemize}
\RecSimNG\ runtimes are also able to handle distributed execution, as well as alternative sampling methods, such as Markov Chain Monte-Carlo samplers. 

Finally, we note that while much of the \RecSimNG\ functionality is currently offered through the TensorFlow back end, the core \RecSimNG\ architecture is back-end independent, enabling applications to be developed within other computational frameworks (such as JAX or PyTorch).

\subsection{Probabilistic Programming}
\label{sec:learning}

We now elaborate on the probabilistic programming ethos that underlies \RecSimNG, which supports sophisticated probabilistic inference, and is critical to model learning and recommender policy optimization.
In Sec.~\ref{sec:overview}, we defined a \RecSimNG\  program as a stochastic process described by DBN. We now discuss in greater details precisely how DBNs are defined in \RecSimNG. Recall that a DBN is a Markovian stochastic process whose kernel factorizes as 
$$T(s_t, s_{t-1}) = \prod_i T^i(s_t^i | s^{\operatorname{Pa}_{\Gamma}(i)}_t, s_{t-1}^{{\operatorname{Pa}_{\Gamma_{-1}}(i)}}),$$ where $\operatorname{Pa}_\Gamma(i)$ and $\operatorname{Pa}_{\Gamma^{-1}}(i)$ denote sets of random variables. respectively, whose current time step (present) and prior time step (previous) values jointly determine the distribution from which $s_t^i$ is generated. 

The equation above embodies a \emph{declarative} style for defining the core stochastic process; that is, to specify a DBN, we specify the conditional distributions $T_i$ using mathematical functions in some symbolic language. This is clearly \emph{not} what we have done in the examples so far, which have used more algorithmic (or \emph{imperative}) definitions by implementing behaviors as sequences of TensorFlow operators which transform given inputs into the desired outputs. 

Generally, imperative specifications tend to be much more intuitive than declarative ones---especially in the context of agent-based modeling, where we can literally imagine an agent engaging in a sequence of steps to accomplish some objective, task or effect. That said, they tend to be less useful than declarative approaches since they only describe how values are produced. As a result, they do not allow us to transform the induced distribution or derive new mathematical facts to aid in specific inference tasks (for example, computing its gradients).

There is, however, a fairly wide set of cases where imperative definitions can be converted to declarative ones in reasonable time and space, allowing us to have the best of both worlds. To achieve this, \RecSimNG\ uses the  probabilistic programming environment Edward2 \cite{tran2018simple}.\footnote{See \texttt{https://github.com/google/edward2}.}
Edward 2 is used as a layer on top of TensorFlow Probability (TFP) which employs program transformation to allow us to treat our imperative sampling code as if it were a symbolic declarative definition of the distribution.\footnote{See \texttt{https://github.com/tensorflow/probability}.}
\RecSimNG\ exposes a small (but easily extensible) set of Edward 2 program transformations tailored to simulation-specific tasks (including those of special relevance for simulating agents involved in recommender systems).

One significant difference between \RecSimNG\ and a pure Monte Carlo simulator is that it can evaluate the probabilities of trajectories according to the DBN induced by the simulation. This functionality is contained in the \inlinecode{log\_probability} module. This module enables the calculation of the log probabilities of data trajectories relative to the stochastic process defined by a \RecSimNG\ model. This, together with the differentiability provided by the TensorFlow runtime, enables the implementation of powerful data assimilation methods. For example, the update step of a maximum-likelihood simulation parameter-fitting method can be as simple as:

\begin{lstlisting}[language=Python]
with tf.GradientTape() as tape:
  log_probs = [
      log_probability.log_probability_from_value_trajectory(recsim_model, traj)
      for traj in trajectories
  ]
  negative_likelihood = -tf.reduce_sum(log_probs)
grad = tape.gradient(negative_likelihood, simulation_parameters)
\end{lstlisting} 

Through the \inlinecode{log\_probability} API, \RecSimNG\ can interface with the MCMC machinery provided by TensorFlow Probability. This powers various posterior inference tasks, which in turn enables model learning given partially observed trajectories. For example, suppose we are given trajectories generated by the user model specified in Sec.~\ref{sec:library}, but where the values of the user interest vectors are not provided---this would reflect the data available to the recommender system itself in most natural settings, which observes only user choice behavior, not the latent state that drives it.
One option for dealing with such missing data (e.g., by the recommender attempting to predict future user choice behavior) would be to use an \emph{expectation-maximization (EM)} \cite{bishop_PRML:2006} algorithm estimate both user choice behavior and its dynamics. An implementation of a stochastic version of EM could be as simple as :

\begin{lstlisting}[language=Python]
for i in range(num_iters):
  posterior_samples = run_chain()
  log_probs = []
  with tf.GradientTape() as tape:
    log_probs = [unnormalized_log_prob_train(posterior_sample) for in posterior_samples]
    neg_elbo = -tf.reduce_mean(log_probs)
  grads = tape.gradient(neg_elbo, [model_parameters])
  optimizer.apply_gradients(zip(grads, [model_parameters])
\end{lstlisting} 

\noindent
Here the \inlinecode{unnormalized\_log\_prob\_train} function injects the proposed interest imputation into the partial trajectory and evaluates its log probability:

\begin{lstlisting}[language=Python]
def unnormalized_log_prob_train(proposed_interest):
  # Hold out the user intent in the trajectories.
  user_state_dict = dict(traj['user state'].as_dict)
  user_state_dict['interest'] = proposed_interest
  traj['user state'] = Value(**user_state_dict)
  return log_probability.log_probability_from_value_trajectory(
      variables=variables, value_trajectory=traj, num_steps=horizon - 1)
\end{lstlisting} 

\noindent
The \inlinecode{run\_chain} function invokes TFP's Hamiltonian Monte Carlo sampler.
This example is developed in full detail in the \RecSimNG\ tutorials.

The \inlinecode{log\_probability} API is highly flexible and, together with differentiability, enables a wide array of training and inference methods, such as: policy gradient reinforcement learning, adversarial training, variational Bayesian inference, generative adversarial networks and many others.

\section{Use Cases}
\label{sec:usecases}

In this section, we offer a small sampling of the ways in which \RecSimNG\ can be used to develop or assess recommender algorithms. We present three use cases---each using a somewhat simplified, stylized model of user or content-provider behavior and relatively simple recommender policies---that showcase the execution and inference capabilities offered by \RecSimNG\ and the specific mechanisms by which it can be used to evaluate, optimize, or train models and algorithms. We leave out many of the details of these models to focus attention on broader uses to which \RecSimNG\ can be put---full details for these use cases (including complete model specifications and parameter values) can be found in the online tutorials.\footnote{See \texttt{https://github.com/google-research/recsim\_ng/tree/master/recsim\_ng/colab}.} We also point to research articles that use more elaborate, realistic models and policies on which several of these use cases are based. (We note that some of the results in this cited work were produced using early versions of \RecSimNG.) Finally, we note that \RecSimNG's TensorFlow-based implementation allows one to train recommender policies through direct coupling with a \RecSimNG\ environment.

\subsection{Partially Observable RL}
\label{sec:usecase_PORL}

One of the greatest challenges facing the deployment of RL in recommender systems is optimizing policies in the presence of user latent state (satisfaction, interests, etc.), unobservable contextual conditions (location, activity, companions), and other latent environmental factors \cite{advamp:ijcai19,dulac-arnold:2019:realworldrl}. The induced partially observable RL problems require defining policies over some summary of the user and/or environment history---effectively involving some form of state estimation---which can be exceptionally challenging.

Fortunately, generally applicable methods like policy-gradient optimization, coupled with variance-reduction heuristics and methods for unbiasing training, often provide good results for non-Markovian problems. However, tuning policy parameters in this way is by its nature heuristic, depending on many factors such as the policy architecture, parameter initialization, choice of optimizer (and its settings), and so on, for which there is very little (general) theoretical guidance. It is hence crucial to understand conditions under which recommender algorithms, and the specific representations they adopt, work well or fail---this can be used to improve algorithm development in research settings, or increase confidence in one's models prior to deploying them in live experiments or production settings. Given a specific user/environment model, \RecSimNG\ makes it quite straightforward to train a policy using policy gradient using TensorFlow, TFP, and its log probability API. We illustrate this using a simplified environment model and policy representation.

We adopt a simple probabilistic factorization model to explore and uncover user interests, and use \RecSimNG\ to learn a latent representation of users using policy gradient. Our simulation environment is similar to that used by Ie, et al.~\cite{slateQarxiv}, but with the addition of partial observability. The recommender policy presents a slate of content items to a user at each interaction, from which the user selects at most one item for consumption. Users and items are modeled in a fashion similar to that described in Sec.~\ref{sec:library}: each user has a dynamic, latent topic/interest vector, whose initial value is sampled randomly, and which evolves based on the topic and the quality of any recommended item consumed. Specifically, if a high-quality item is consumed, a user's interests move (stochastically) in the direction of the item's topic, while lower-quality consumption has the opposite effect. The quality of an item also impacts the user's immediate engagement with an item: lower quality items have lower expected engagement. Finally, the probability of a user selecting (and engaging) with a recommended item uses a simple MNL model: the selection probability is greater if that item's topic is closer to the user's \emph{current} interest vector.

The \RecSimNG\ environment also correlates topics with quality: the quality distribution for some topics skews higher than for others. Thus, the recommender system faces interesting short/long-term tradeoffs, as well as exploration/exploitation tradeoffs, of the type familiar to RL researchers and practitioners: (a) to promote greater user engagement over the long-run, the recommender is inclined to recommend items from high-quality topics, which will gradually nudge the user's interests toward such topics; (b) however, such a strategy comes at a short-term cost if the recommended items are not close to the user's \emph{current} interests; and (c) the optimal items(s) to place on a slate to make this tradeoff depend on the user's latent interest vector, the estimation of which the recommender can improve by taking active exploration or probing steps by recommending specific items.

To reflect the partially observable nature of the problem, we design a recommender \emph{policy class} that maintains a finite history of its engagement with a user consisting of the past $15$ consumed items and learns an embedding of this history. Based on this embedding and user engagement, the recommender uses a deep network to learn a \emph{belief state} (or estimate of the user latent state) $h_u$; it also learns a similar item representation. Finally, we train the stochastic recommender policy (which selects a slate of items based on its belief state for the user in question) using the policy gradient method REINFORCE~\cite{Williams92simplestatistical}, where the policy parameters are comprised of the embeddings and the parameters that encode $h_u$.  Specifically, using \RecSimNG, given the current policy, we sample trajectories for $B$ users over horizon $T$, and then iteratively update the policy parameters using the policy gradients (of the score function surrogate of the cumulative user engagement) computed by automatic differentiation in TensorFlow. In the snippet below, the \inlinecode{slate docs log prob} variable, which computes the log probability of the slates composed by the recommender, is automatically generated by \RecSimNG's log probability API.

\begin{lstlisting}[language=Python]
with tf.GradientTape() as tape:
    last_state = tf_runtime.execute(num_steps=horizon - 1)
    last_metric_value = last_state['metrics state'].get('cumulative_reward')
    log_prob = last_state['slate docs log prob'].get('doc_ranks')
    objective = -tf.tensordot(tf.stop_gradient(last_metric_value), log_prob, 1)
    objective /= float(num_trajectories)

  grads = tape.gradient(objective, trainable_variables)
  if optimizer:
    grads_and_vars = list(zip(grads, trainable_variables))
    optimizer.apply_gradients(grads_and_vars)
\end{lstlisting}

\begin{figure*}[t]
  \centering
  \includegraphics[width=2.8in]{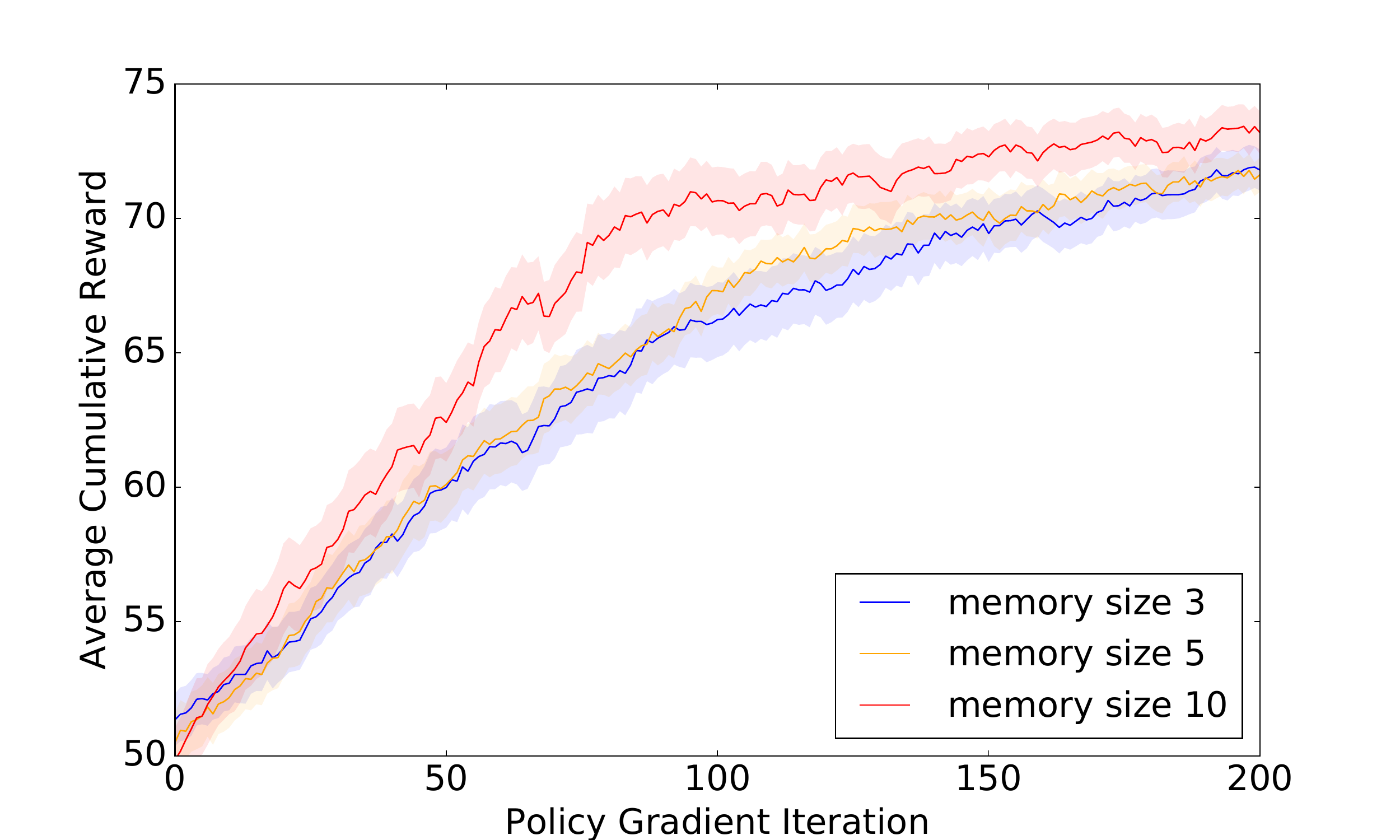}
  \vspace{-0.1in}
  \caption{The cumulative consumption time as a function of REINFORCE iterations averaged over 20 runs. Results shown for different amounts of history (memory) used as input to the policy.}
  \label{fig:policy_gradient}
\end{figure*}

Fig.~\ref{fig:policy_gradient} shows how, in a simple experiment, average-per-user cumulative engagement time induced by the REINFORCE-trained policy improves with the number of policy-gradient iterations.\footnote{The size of the slate is two and $d$ is twenty. The number of users $B$ in each batch is $1000$ and the horizon $T$ is 100.}
This demonstrates how \RecSimNG\ can be used to power policy-gradient training to solve partially-observable RL problems. We note that the  policy class over which we optimize is intentionally very simple for illustrative purposes. But the same methodology can be applied to more realistic models and policy classes, e.g., RNN-based policies---indeed, this technique has been applied to  contextual bandit problems, where \RecSimNG\ is used in the \emph{meta-learning} of problem-distribution-specific bandit policies that minimize Bayes regret \cite{diff_bandit:nips20}. 

\RecSimNG\ makes it very easy to compare different policy classes or learning methods under a variety of different model conditions. For instance, we can change the amount of user history used by the policy, use REINFORCE to optimize the resultant policies and measure the impact on performance: Fig.~\ref{fig:policy_gradient} also includes training performance curves for several different history lengths. It is also straightforward to, for instance: (a) change the degree to which user interests evolve---from purely static to very rapid---and test how much history is required to get good policy performance under different conditions; (b) vary the optimization horizon to compare the performance of myopic policies to RL-based policies; or (c) vary the user choice model to study the impact on the resulting optimized policies.

\subsection{Learning Latent Variable Models}
\label{sec:usecase_latentvar}

The use case in Sec.~\ref{sec:usecase_PORL} shows the importance of managing partial observability in the design and optimization of recommender policies. There, partial observability---and hence the non-stationarity of the dynamics relative to the features observable by the recommender---was driven by a user's \emph{latent (and evolving) interests}. Note however that the recommender made no assumptions about the source of this non-stationarity. By contrast, a policy based on a model which makes this latent-state structure explicit can often be easier to learn/optimize.

More generally, learning \emph{latent variable models} can be extremely useful for recommender systems, as they not only allow for the effective compression of interaction histories or activity streams, but also offer the prospect of greater interpretability through the lens of meaningful attributes such as user interests, intent, satisfaction, so on. Of course, learning latent variable models relies in fundamental ways on probabilistic inference---in particular, the computation of the posterior of the latent variable(s) given observation histories. Such inference is readily enabled by \RecSimNG's APIs, and can be tightly integrated with TFP's inference machinery, which we illustrate using one of its MCMC routines.

We demonstrate the use of \RecSimNG\ to learn a simple latent variable model by generating data streams capturing the interaction of users with a recommender whose policy is ``fixed'' (i.e., non-adaptive). The user-recommender interaction is like that in the previous use case---a user is presented with a slate of recommended items and chooses one (or zero) for consumption using an MNL choice model, based on affinity with the user's interest vector. The recommender policy is a non-adaptive, randomized policy. More critically, the user model differs from the one above in two ways. First, each user's (latent) interest vector is \emph{static}. Second, each user has a dynamic, scalar \emph{satisfaction} level that increases or decreases depending on immediate ``trend'' in recommendation quality. More precisely, if the most attractive item on the recommended slate at time $t$ is worse that the most attractive item on the previous slate at time $t-1$ (i.e., if the user has less affinity, given her latent interest vector, with the current item than the prior item), then the user's satisfaction (stochastically) decreases. Likewise, if the most attractive item on the slate improves from one stage to the next, satisfaction (stochastically) increases. The amount of increase or decrease at each interaction is governed by a latent \emph{sensitivity} parameter $\alpha_u$ which differs for each user $u$---specifically, $\alpha_u$ influences the mean of a normal distribution from which the updated satisfaction is drawn.\footnote{We refer to the online tutorial for a complete specification of the model.}
The user's choice process either selects a item for consumption, using the sum of an item's affinity (computed as the negative Euclidean distance between the item and the intent) and the user's satisfaction as logits, or abstains according to a constant ``no choice'' logit (set to 0 for illustration). The user's satisfaction effectively acts as a boost to all item logits compared to the constant "no choice" logit; thus, at high levels of satisfaction, the user is more likely to select some item for consumption rather than opt out with ``no choice.'' If the user's satisfaction drops sufficiently and no good items are recommended, the user effectively drops out of the recommender platform in the sense that the probability of selecting any item becomes negligible.

\begin{figure*}[t]
  \centering
  \includegraphics[width=3.2in]{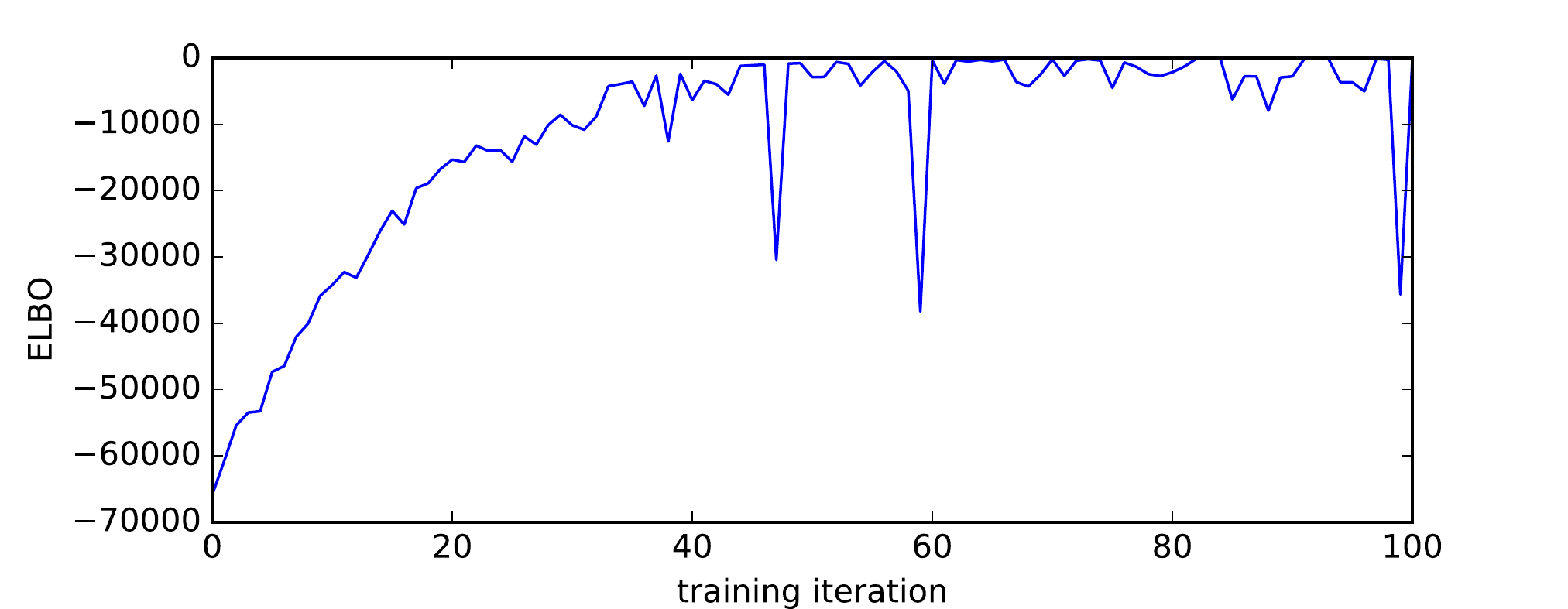}
  \vspace{-0.1in}
  \caption{ELBO versus training step of the Monte-Carlo EM algorithm.}
  \label{fig:mcem}
\end{figure*}

We test \RecSimNG's model learning capabilities by treating user satisfaction as observable (e.g., it may correspond to a user's observable degree of engagement with a consumed), but attempt to learn each user's latent interest vector and sensitivity parameter $\alpha_u$.
We use the \RecSim\ model above (the ground-truth model) to generate user trajectories, i.e., training data.
We use a \emph{different} \RecSimNG\ model (the learning model) to train using this trajectory data: it's goal is to fit a new user model to the generated ``ground-truth'' trajectory data. 
In particular, we use a simple Hamiltonian Monte Carlo~\cite{neal2011hmc} routine implemented in TensorFlow Probability as a posterior sampler within a Monte-Carlo EM learning algorithm (as outlined in Sec~\ref{sec:learning}).
In the learned simulation model, initially $\alpha_u$ is sampled from $U(0, 1)$. In this example, the learning algorithm recovers the ground-truth model parameters. The progress of the algorithm is illustrated in Fig.~\ref{fig:mcem} by means of the increasing 
\emph{evidence lower bound (ELBO)} over training iterations. 

\subsection{Ecosystem Modeling}
\label{sec:usecase_ecosys}

One aspect of recommender systems research that has received relatively little attention is the role of multiagent interactions among participants (e.g, different users, content providers, etc.) in the recommender ecosystem, and the impact these have on recommendation policy quality.\footnote{There are exceptions, of course, for example, recent work on fairly  allocating the user exposure to content providers \cite{SinghJoachims2018,Biega}, game-theoretic optimization of policies when providers act strategically \cite{benporat_etal:nips18,benporat_etal:aaai19}, or optimizing collective user utility (or social welfare) over extended horizons when recommender policies induce dynamic behavior among providers \cite{mladenov_etal:icml20}. We adopt a simplified variant of this latter model in this section.}
Using simulation to reason about recommender ecosystems---and optimize recommender policies---requires a flexible modeling framework that supports the effective expression of the behaviors of all participants. Moreover, since the behavior trajectories of the ecosystem agents are no longer independent---e.g., the behavior of a user might influence the response of a content provider, or the recommender policy may correlate the behaviors of multiple users---simulation is no longer ``embarrassingly parallel.'' In such cases, it is critical that simulation libraries offer superior performance, especially when the number of agents is large. \RecSimNG\ offers reusable modeling blocks---for example, the same types of state and choice models can be used for both content provider and users/content consumers---and can enhance performance considerably using graph compilation, Tensorflow XLA (accelerated linear algebra) and TPU accelerators. 

To demonstrate these capabilities, we develop a simple ecosystem model in which a recommender system mediates the interaction between users and content providers by making recommendations to users over time. This model is a (very simplified) version of that developed by Mladenov \emph{et al.}~\cite{mladenov_etal:icml20}, who study the long-term equilibria of such ecosystems. We adopt a simplified user model whereby each user is characterized by a static, observable user interest vector. As above, the vector determines item affinities, which in turn are used as inputs to a multinomial logit choice model that determines item selection from a recommended slate. User utility for any consumed item is simply her affinity for the item perturbed with Gaussian noise. The aim of the recommender is to maximize cumulative user utility (over all users) over a fixed horizon.

Ecosystem effects emerge because of content provider behavior. Every recommendable item is made available by a unique provider $c$. Like users, each provider has an ``interest vector'' around which the items it makes available are centered. User and provider interest vectors are sampled using a mixture model to generate subpopulations or ``communities'' of different sizes, with members of any given community having interests that are more-or-less aligned.
Moreover, each provider $c$ is assumed to require a certain amount of \emph{cumulative past-discounted engagement} $E_c$ over any fixed period: here engagement is measured by the cumulative number of its content items consumed by any user over the recent past, with more distant consumption discounted relative to more recent consumption. The greater $E_c$ is, the greater the number of items (randomly drawn near its interest vector) that provider $c$ will generate at any given period.

We compare two different recommender policies in this setting. The first is a standard ``myopic'' policy that, for any user $u$, always recommends the slate of $k$ items for which the user has greatest affinity (we assume that, somewhat unrealistically, the recommender can observe the user's interest vector, in order to simplify the model to focus on ecosystem interactions). Under such a policy, the behavior of providers has the potential to give rise to ``rich-get-richer'' phenomena: providers that initially (e.g., for random reasons) attract significant engagement make available a \emph{greater number of items} at subsequent periods, which increases the odds of attracting even further future engagement; by contrast, providers with limited engagement tend to reduce the number of items they generate, which can decrease future engagement. These dynamics can lead to the eventual decrease in diversity of the type of content made available, with a concomitant decrease in overall user utility (specifically, disadvantaging users whose interests lie near those of low-engaged providers).

The second recommender policy is ``aware'' of these provider dynamics, which it counteracts by generating recommended slates using a content provider ``boost'' $B_c$ to score the top $k$ items to add to the recommended slate. Specifically, the boost $B_c$ for provider $c$ increases (decreases) proportionally to the difference between $c$'s current cumulative discounted engagement and the mean engagement across all providers. This boost (which is capped in a tunable fashion via an auxiliary parameter $L$) is then used to randomly increase (or decrease if it is negative) each item's affinity for a given user $u$. This increases the odds of low-engaged providers having their content recommended on the slate of top $k$ items.\footnote{We note that the ecosystem-aware policy here uses a very simple heuristic, and refer to \cite{mladenov_etal:icml20} for a formal analysis and more sophisticated policy optimization algorithm for this type of problem.}

\begin{figure*}[t]
  \centering
  \includegraphics[width=2.8in]{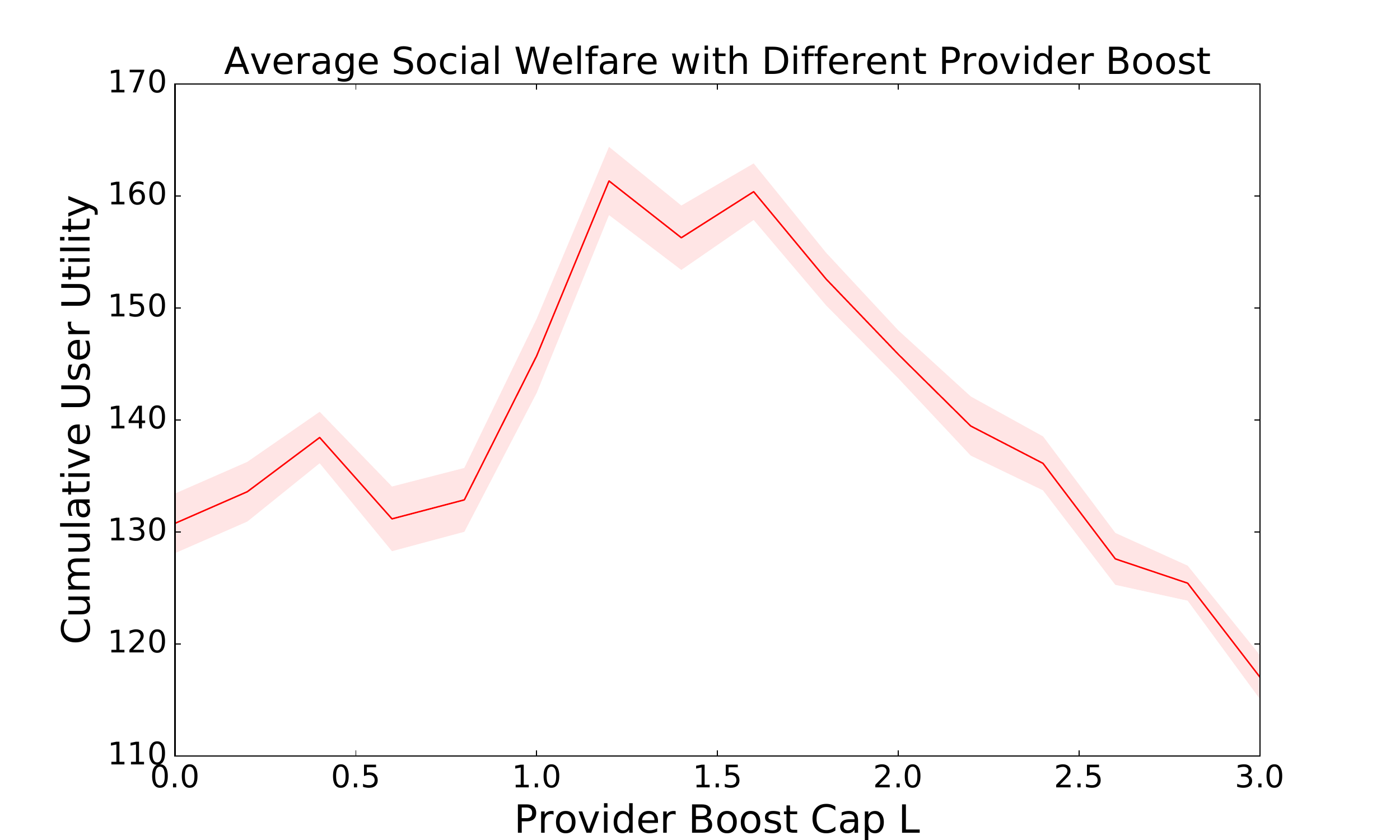}
  \vspace{-0.1in}
  \caption{Cumulative user utility averaged over all users, for various values of the boost cap parameter $L$ (40 runs). The shaded region shows standard error.}
  \label{fig:ecosystem}
\end{figure*}

We use \RecSimNG\ to compare the two policies with respect to long-term average user utility (i.e.,  user social welfare) by randomly sampling 2000 users, 80 providers, 400 items per period and measuring cumulative user utility over a horizon of 300 periods. In fact, we evaluate several different instantiations of ecosystem-aware policy for different values of the boost-cap parameter $L$: here $L=0$ corresponds to a myopic policy (it offers no boost at all), while very large values of $L$ promote under-served providers in a very aggressive way. Fig~\ref{fig:ecosystem} plots the cumulative user utility (averaged over $40$ random runs) and shows that an ecosystem-aware policy with $L=1.2$ achieves significantly better cumulative user utility ($161.34$) than the myopic policy ($130.77$); but as this parameter is increased further, cumulative utility decreases due to the recommendation of content that is increasingly far away from the user's interests. \RecSimNG\ can be used to explore the impact of different policy formats, parametrizations or objective functions easily; but it can also be used to assess under which environment conditions different policies work well. For example, in this model, it is straightforward to change the underlying (random) ``community structure'' by adjusting the joint distribution from which user and provider interests are sampled.\footnote{As an example, the model used here induces community structure hierarchically by first sampling provider interests, then sampling user interests by treating provider interests as the core of a mixture model (see the online tutorials for details), the parameters of which can be used to generate ``tighter'' or ``looser'' communities (users with similar interests) of varying size.} Likewise, one can adjust the user choice model, the provider content-generation model, etc.

Finally, we note that many simulation runs are typically required to reduce confidence intervals of around the estimates of random variables of interest (e.g., average user utility) to acceptable levels. In such circumstances, performance characteristics are critical. We compared the runtime of this ecosystem simulation on CPU vs.\ Google TPUv2, while varying the number of simulation runs from 8 to 1024 (each run is a sampled ecosystem). 
We found that the inherent parallelization capabilities of \RecSimNG\ allow Google TPUv2 to achieve roughly a 6X speedup across a variety of user population sizes.

\section{Concluding Remarks}

We have outlined \RecSimNG, a  scalable, modular, differentiable, probabilistic platform for the simulation of multi-agent recommender systems. Implemented in Edward2 and TensorFlow, it offers: a powerful, general probabilistic programming language for agent-behavior specification; tools for probabilistic inference and latent-variable model learning, backed by automatic differentiation and tracing; and a TensorFlow-based runtime for executing simulations on accelerated hardware. We illustrated how \RecSimNG\ can be used to create transparent, configurable, end-to-end models of a recommender ecosystem. We also outlined a set of simple use cases that demonstrated how \RecSimNG\ can be used to: develop and evaluate interactive, sequential RL-based recommenders algorithms; learn latent-state models from data trajectories; and evaluate different recommender policies in complex multi-agent domains with substantial agent interaction. We also pointed to other research efforts that used \RecSimNG\ to evaluate more sophisticated algorithms (beyond the simple methods we used for illustration purposes in this paper). 

Our hope is that \RecSimNG\ will  help both researchers and practitioners easily develop, train and evaluate novel algorithms for recommender systems. There are a number of profitable next steps we plan to take to further this objective. Obviously, the development and distribution of models that cover additional use cases will help exemplify new design patterns and algorithmic principles for those looking to avail themselves of \RecSimNG. As we continue to apply \RecSimNG\ to novel problems, we plan to add new, reusable modeling components to the open-source library---we also welcome and encourage contributions to the open-source library from other developers and users of \RecSimNG. We are also investigating the release of increasingly realistic user models that can serve as benchmarks for the research community, as well as methods that can facilitate ``sim-to-real'' transfer using \RecSimNG.

\bibliographystyle{ACM-Reference-Format}

\end{document}